\documentclass[11pt,a4paper]{article}
\usepackage[toc,page]{appendix}
\usepackage{authblk}
\usepackage[hyperref]{acl2018}
\usepackage{times}
\usepackage{latexsym}
\usepackage{graphicx}
\usepackage{enumitem}  
\usepackage{caption}
\usepackage{url}
\usepackage{amsmath}
\usepackage{nicefrac}
\usepackage{enumitem}
\usepackage{subcaption}
\usepackage{multirow}
\usepackage{float}
\usepackage{mdframed}
\usepackage{soul}

\aclfinalcopy
\def\aclpaperid{1005}

\title{Unsupervised Abstractive Meeting Summarization with Multi-Sentence Compression and Budgeted Submodular Maximization\Thanks{ Published as a long paper at ACL 2018. $^\dagger$Work done as part of 3$\mathrm{^{rd}}$ year project, with equal contribution.}}

\author[1,2]{Guokan Shang}
\author[1]{Wensi Ding$^\dagger$}
\author[1]{Zekun Zhang$^\dagger$} 
\author[1]{Antoine J.-P. Tixier}
\author[1,3]{\\Polykarpos Meladianos}
\author[1,3]{Michalis Vazirgiannis}
\author[2]{Jean-Pierre Lorr\'e}
\affil[1]{\'Ecole Polytechnique, $^\mathrm{2}$Linagora, $^\mathrm{3}$AUEB}

\date{\today}

\begin{document}
\maketitle
\begin{abstract} 
We introduce a novel graph-based framework for abstractive meeting speech summarization that is fully unsupervised and does not rely on any annotations. Our work combines the strengths of multiple recent approaches while addressing their weaknesses. 
Moreover, we leverage recent advances in word embeddings and graph degeneracy applied to NLP to take exterior semantic knowledge into account, and to design custom diversity and informativeness measures. Experiments on the AMI and ICSI  corpus show that our system improves on the state-of-the-art. Code and data are publicly available\footnote{\tiny{\url{https://bitbucket.org/dascim/acl2018_abssumm}}}, and our system can be interactively tested\footnote{\tiny{\url{http://datascience.open-paas.org/abs_summ_app}}}.
\end{abstract}

\section{Introduction}
People spend a lot of their time in meetings. The ubiquity of web-based meeting tools and the rapid improvement and adoption of Automatic Speech Recognition (ASR) is creating pressing needs for effective meeting speech summarization mechanisms.

Spontaneous multi-party meeting speech transcriptions widely differ from traditional documents. Instead of grammatical, well-segmented \textit{sentences}, the input is made of often ill-formed and ungrammatical text fragments called \textit{utterances}. On top of that, ASR transcription and segmentation errors inject additional noise into the input.

In this paper, we combine the strengths of 6 approaches that had previously been applied to 3 different tasks (keyword extraction, multi-sentence compression, and summarization) into a unified, fully unsupervised end-to-end meeting speech summarization framework that can generate readable summaries despite the noise inherent to ASR transcriptions. We also introduce some novel components. Our method reaches state-of-the-art performance and can be applied to languages other than English in an almost out-of-the-box fashion.

\section{Framework Overview}

\begin{figure*}[ht]
\centering
\includegraphics[scale=0.55]{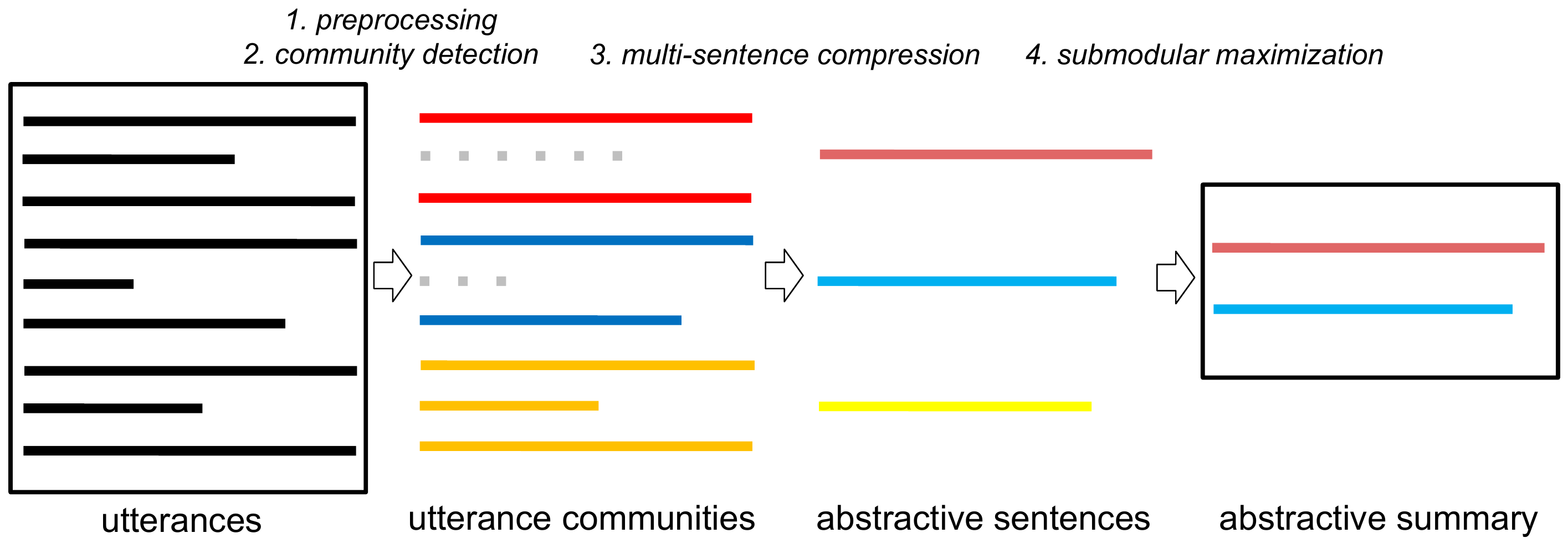}
\caption{Overarching system pipeline.}
\label{fig:architecture}
\end{figure*}

As illustrated in Figure \ref{fig:architecture}, our system is made of 4 modules, briefly described in what follows.  

The first module pre-processes text. The goal of the second \textit{Community Detection} step is to group together the utterances that should be summarized by a common abstractive sentence \cite{murray2012using}. These utterances typically correspond to a topic or subtopic discussed during the meeting. A single abstractive sentence is then separately generated for each community, using an extension of the Multi-Sentence Compression Graph (MSCG) of Filippova \shortcite{filippova2010multi}. Finally, we generate a summary by selecting the best elements from the set of abstractive sentences under a budget constraint. We cast this problem as the maximization of a custom submodular quality function.

Note that our approach is fully unsupervised and does not rely on any annotations. Our input simply consists in a list of utterances without any metadata. All we need in addition to that is a part-of-speech tagger, a language model, a set of pre-trained word vectors, a list of stopwords and fillerwords, and optionally, access to a lexical database such as WordNet. Our system can work out-of-the-box in most languages for which such resources are available.

\section{Related Work and Contributions} \label{sec:related_work}
As detailed below, our framework combines the strengths of 6 recent works. It also includes novel components.

\subsection{Multi-Sentence Compression Graph (MSCG) \cite{filippova2010multi}}

\underline{Description}: a fully unsupervised, simple approach for generating a short, self-sufficient sentence from a cluster of related, overlapping sentences. As shown in Figure \ref{fig:msc_illustration}, a word graph is constructed with special edge weights, the $K$-shortest weighted paths are then found and re-ranked with a scoring function, and the best path is used as the compression. The assumption is that redundancy alone is enough to ensure informativeness and grammaticality.\\
\underline{Limitations}: despite making great strides and showing promising results, Filippova \shortcite{filippova2010multi} reported that 48\% and 36\% of the generated sentences were missing important information and were not perfectly grammatical.\\
\underline{Contributions}: to respectively improve informativeness and grammaticality, we combine ideas found in Boudin and Morin \shortcite{boudin2013keyphrase} and Mehdad et al. \shortcite{mehdad2013abstractive}, as described next.

\subsection{More informative MSCG \cite{boudin2013keyphrase}}
\underline{Description}: same task and approach as in Filippova \shortcite{filippova2010multi}, except that a word co-occurrence network is built from the cluster of sentences, and that the PageRank scores of the nodes are computed in the manner of Mihalcea and Tarau \shortcite{mihalcea2004textrank}. The scores are then injected into the path re-ranking function to favor informative paths.\\
\underline{Limitations}: PageRank is not state-of-the-art in capturing the importance of words in a document. Grammaticality is not considered. \\
\underline{Contributions}: we take grammaticality into account as explained in subsection \ref{sub:mehdad}. We also follow recent evidence \cite{tixier2016graph} that \textit{spreading influence}, as captured by graph degeneracy-based measures, is better correlated with ``keywordedness'' than PageRank scores, as explained in the next subsection.

\subsection{Graph-based word importance scoring \cite{tixier2016graph}}\label{app:a}
\begin{figure}[ht]
\centering
\captionsetup{size=small}
\includegraphics[scale=0.32]{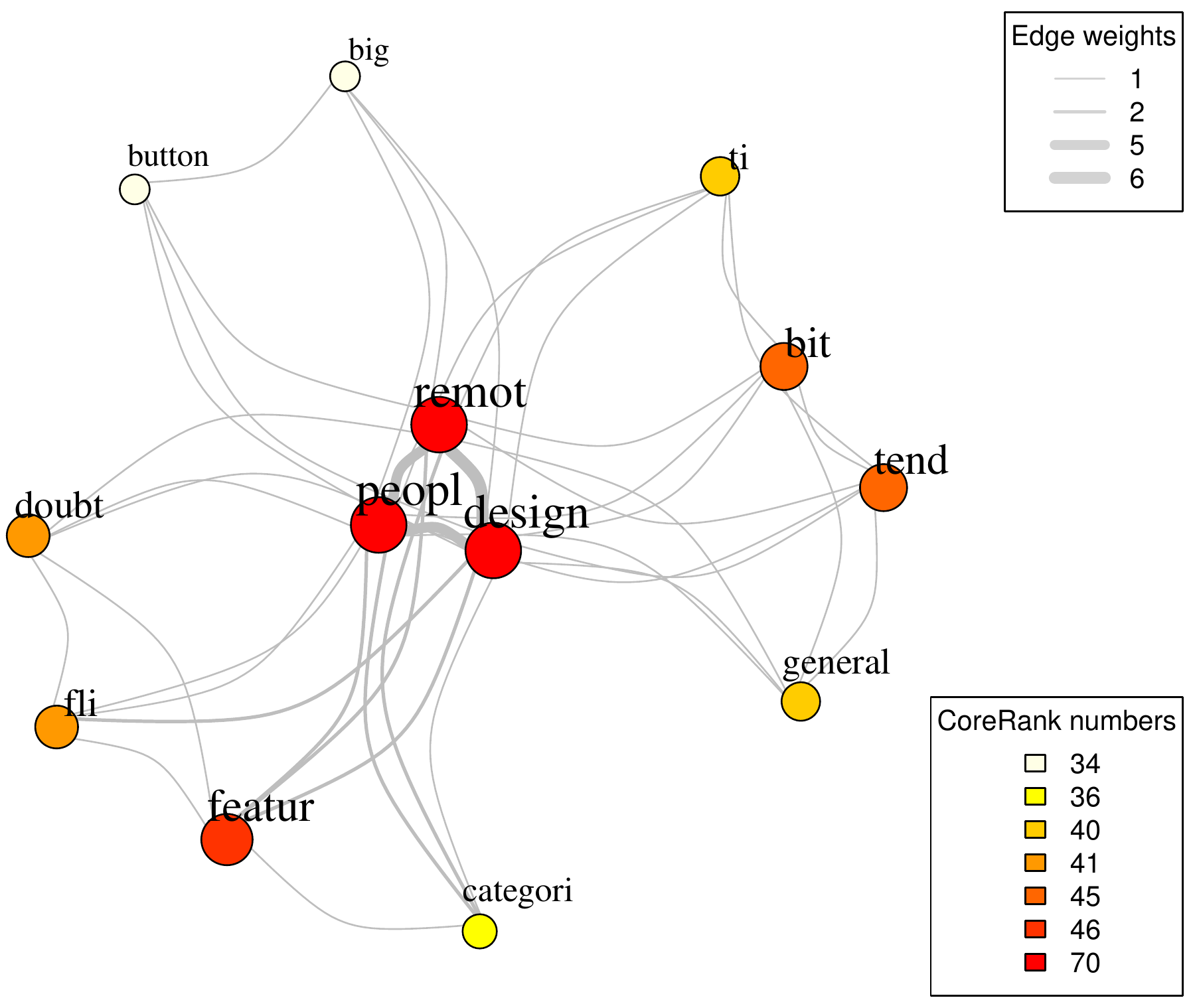}
\caption{
Word co-occurrence graph example, for the input text shown in Figure \ref{fig:msc_illustration}.
}
\label{fig:gow_illustration}
\end{figure}

\noindent \textbf{Word co-occurrence network}. As shown in Figure \ref{fig:gow_illustration}, we consider a word co-occurrence network as an undirected, weighted graph constructed by sliding a fixed-size window over text, and where edge weights represent co-occurrence counts \cite{tixier2016gowvis,mihalcea2004textrank}.

\begin{figure}[ht]
    \centering
    \captionsetup{size=small}
    \includegraphics[scale=0.71]{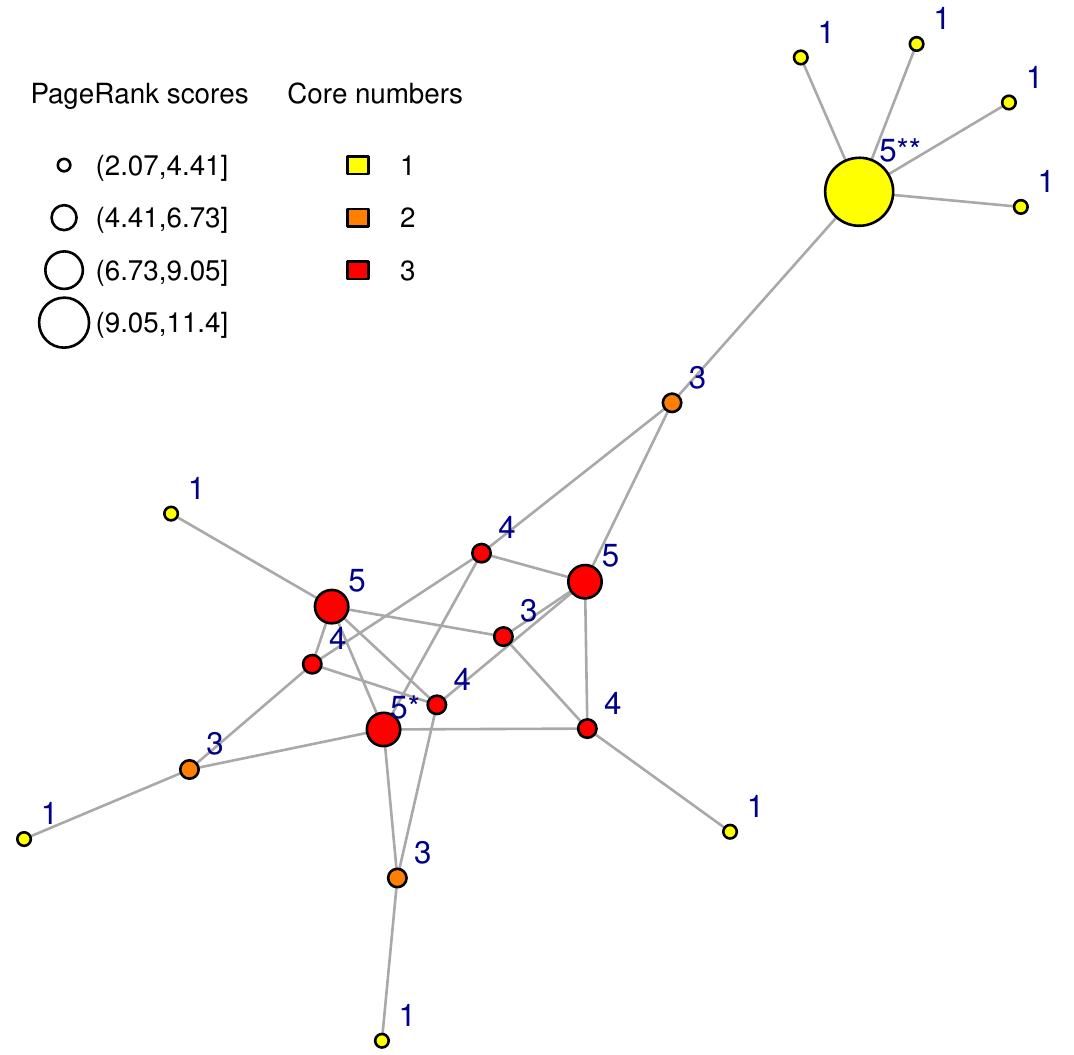}
	\caption{$k$-core decomposition. The nodes $\star$ and $\star\star$ have same degree and similar PageRank numbers. However, node $\star$ is a much more influential spreader as it is strategically placed in the core of the network, as captured by its higher core number.\label{fig:deg_vs_core}}
    \label{fig:kcore}
\end{figure}

\noindent \textbf{Important words are influential nodes}. In social networks, it was shown that \textit{influential spreaders}, that is, those individuals that can reach the largest part of the network in a given number of steps,~are better identified via their core numbers rather than via their PageRank scores or degrees \cite{kitsak2010identification}. See Figure \ref{fig:deg_vs_core} for the intuition. Similarly, in NLP, Tixier et al. \shortcite{tixier2016graph} have shown that keywords are better identified via their core numbers rather than via their TextRank scores, that is, keywords are \textit{influencers} within their word co-occurrence network.

\begin{figure}[t]
\centering
\includegraphics[width=0.37\textwidth]{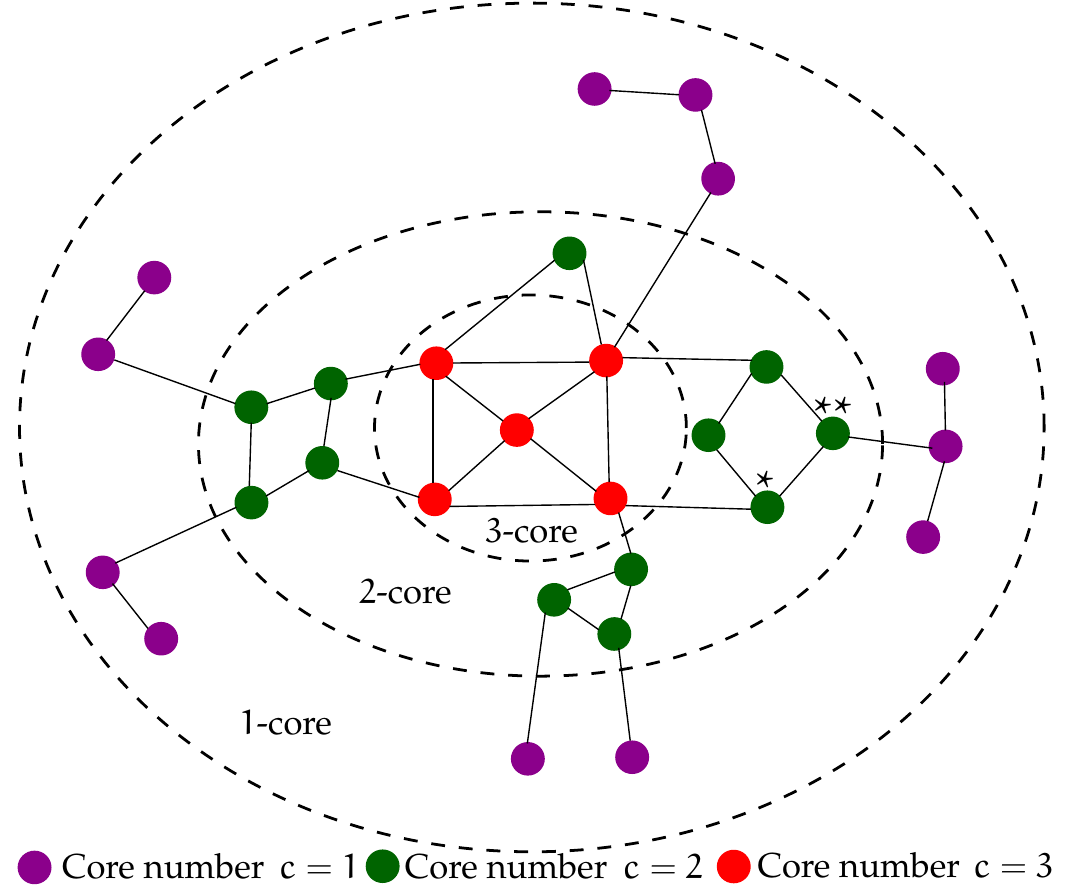}
\captionsetup{size=small}
\caption{Value added by CoreRank: while nodes $\star$ and $\star\star$ have the same core number (=2), node $\star$ has a greater CoreRank score (3+2+2=7 vs 2+2+1=5), which better reflects its more central position in the graph.}
\label{fig:core_rank_illustration}
\end{figure}

\noindent \textbf{Graph degeneracy} \cite{seidman1983network}. 
Let $G(V,E)$ be an undirected, weighted graph with $n=\left\vert{V}\right\vert$ nodes and $m= \left\vert{E}\right\vert$ edges. A $k$-core of $G$ is a maximal subgraph of $G$ in which every vertex $v$ has at least weighted degree $k$. As shown in Figures \ref{fig:kcore} and \ref{fig:core_rank_illustration}, the $k$-core decomposition of $G$ forms a hierarchy of nested subgraphs whose cohesiveness and size respectively increase and decrease with $k$. The higher-level cores can be viewed as a \textit{filtered version} of the graph that excludes noise. This property is highly valuable when dealing with graphs constructed from noisy text, like utterances.
The core number of a node is the highest order of a core that contains this node.

The CoreRank number of a node \cite{tixier2016graph,bae2014identifying} is defined as the sum of the core numbers of its neighbors. As shown in Figure \ref{fig:core_rank_illustration}, CoreRank more finely captures the structural position of each node in the graph than raw core numbers. Also, stabilizing scores across node neighborhoods enhances the inherent noise robustness property of graph degeneracy, which is desirable when working with noisy speech-to-text output.

\noindent \textbf{Time complexity}. Building a graph-of-words is $\mathcal{O}(nW)$, and computing the weighted $k$-core decomposition of a graph requires $\mathcal{O}(m \log(n))$ \cite{batagelj2002}. For small pieces of text, this two step process is so affordable that it can be used in real-time \cite{meladianos2017real}. Finally, computing CoreRank scores can be done with only a small overhead of $\mathcal{O}(n)$, provided that the graph is stored as a hash of adjacency lists. Getting the CoreRank numbers from scratch for a community of utterances is therefore very fast, especially since typically in this context, $n \sim 10$ and $m \sim 100$.

\subsection{Fluency-aware, more abstractive MSCG \cite{mehdad2013abstractive}}\label{sub:mehdad}
\underline{Description}: a \textit{supervised} end-to-end framework for abstractive meeting summarization. Community Detection is performed by (1) building an utterance graph with a logistic regression classifier, and (2) applying the CONGA algorithm. Then, before performing sentence compression with the MSCG, the authors also (3) build an entailment graph with a SVM classifier in order to eliminate redundant and less informative utterances. In addition, the authors propose the use of WordNet \cite{miller1995wordnet} during the MSCG building phase to capture lexical knowledge between words and thus generate more abstractive compressions, and of a language model when re-ranking the shortest paths, to favor fluent compressions.

\noindent \underline{Limitations}: this effort was a significant advance, as it was the first application of the MSCG to the meeting summarization task, to the best of our knowledge. However, steps (1) and (3) above are complex, based on handcrafted features, and respectively require annotated training data in the form of links between human-written abstractive sentences and original utterances and multiple external datasets (e.g., from the Recognizing Textual Entailment Challenge). Such annotations are costly to obtain and very seldom available in practice.\\
\underline{Contributions}: while we retain the use of WordNet and of a language model, we show that, without deteriorating the quality of the results, steps (1) and (2) above (Community Detection) can be performed in a much more simple, completely unsupervised way, and that step (3) can be removed. That is, the MSCG is powerful enough to remove redundancy and ensure informativeness, should proper edge weights and path re-ranking function be used.\\

\vspace{-.3cm}

In addition to the aforementioned contributions, we also introduce the following novel components into our abstractive summarization pipeline:

$\bullet$ we inject global exterior knowledge into the edge weights of the MSCG, by using the \textit{Word Attraction Force} of Wang et al. \shortcite{wang2014corpus}, based on distance in the word embedding space,

$\bullet$ we add a diversity term to the path re-ranking function, that measures how many unique clusters in the embedding space are visited by each path,

$\bullet$ rather than using all the abstractive sentences as the final summary like in Mehdad et al. \shortcite{mehdad2013abstractive}, we maximize a custom submodular function to select a subset of abstractive sentences that is near-optimal given a budget constraint (summary size). A brief background of submodularity in the context of summarization is provided next.

\subsection{Submodularity for summarization \cite{lin2010multi,lin2012submodularity}}\label{sub:rel_sub}
Selecting an optimal subset of abstractive sentences from a larger set can be framed as a budgeted submodular maximization task:

\begin{equation}
\operatorname*{argmax}_{S \subseteq \mathcal{S}} f(S) | \sum_{s \in S} c_s\leq \mathcal{B}
\end{equation}
where $S$ is a summary, $c_s$ is the cost (word count) of sentence $s$, $\mathcal{B}$ is the desired summary size in words (budget), and $f$ is a summary quality scoring set function, which assigns a single numeric score to a summary $S$.

This combinatorial optimization task is NP-hard. However, near-optimal performance can be guaranteed with a modified greedy algorithm \cite{lin2010multi} that iteratively selects the sentence $s$ that maximizes the ratio of quality function gain to scaled cost $\nicefrac{f(S \cup s)-f(S)}{c^r_s}$ (where $S$ is the current summary and $r \geq 0$ is a scaling factor).

In order for the performance guarantees to hold however, $f$ has to be \textit{submodular} and \textit{monotone non-decreasing}. Our proposed $f$ is described in subsection \ref{subsec:submodularity}.

\section{Our Framework} \label{sec:proposed_system}

\begin{figure*}[!ht]
\centering
\captionsetup{size=small}
\begin{minipage}[c]{0.34\textwidth}
\caption{
Compressed sentence (in \textbf{\color{red}{bold red}}) generated by our multi-sentence compression graph (MSCG) for a 3-utterance community from meeting IS1009b of the AMI corpus. Using Filippova \shortcite{filippova2010multi}'s weighting and re-ranking scheme here would have selected another path: \textit{design different remotes for different people bit of it's from their tend to for ti}. Note that the compressed sentence does not appear in the initial set of utterances, and is compact and grammatical, despite the redundancy, transcription and segmentation errors of the input. The \textit{abstractive} and \textit{robust} nature of the MSCG makes it particularly well-suited to the meeting domain.
} \label{fig:msc_illustration}
\end{minipage}\hfill
\begin{minipage}[c]{0.6\textwidth}
\includegraphics[scale=0.57]{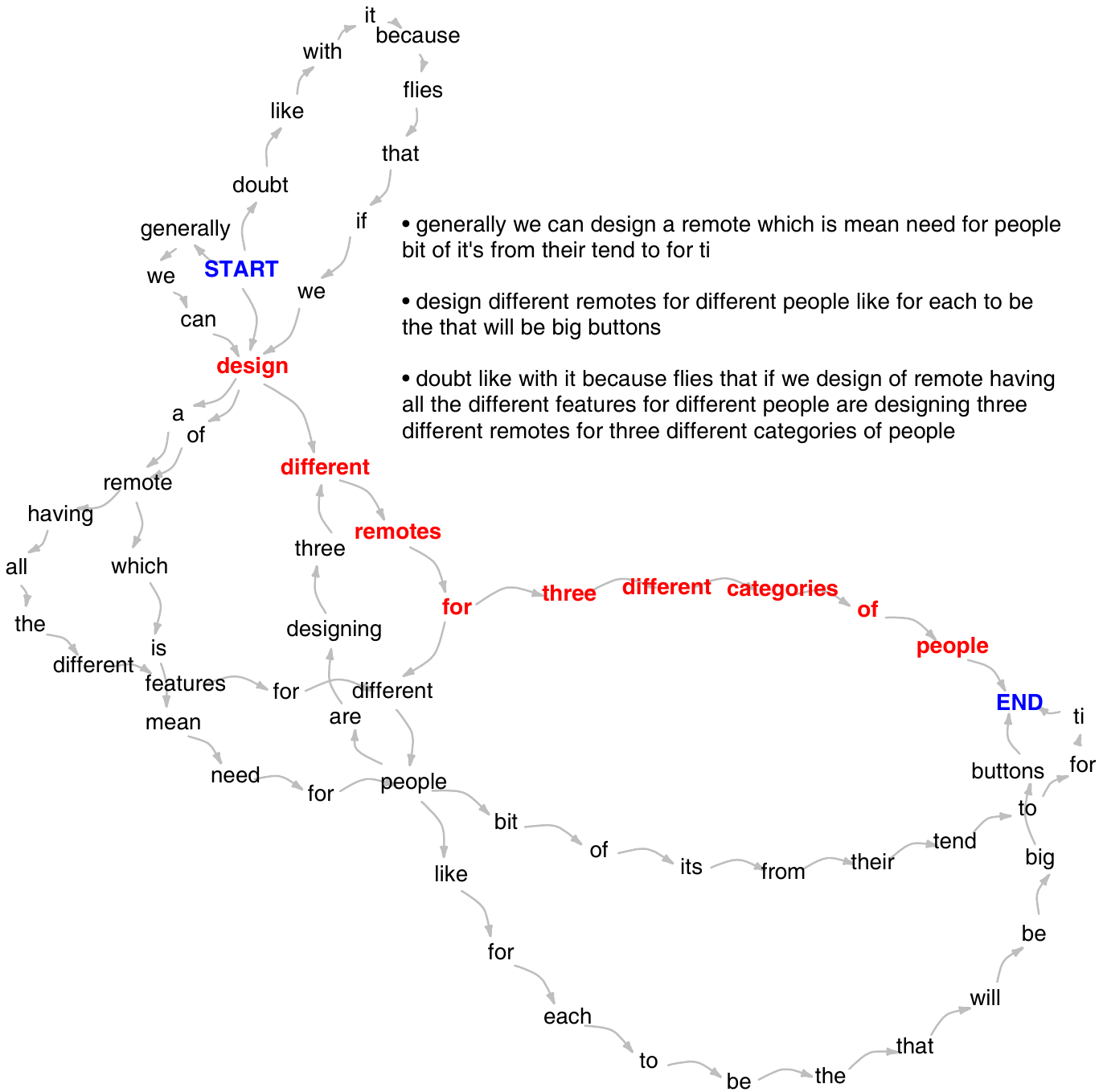}
\end{minipage}

\end{figure*}

We detail next each of the four modules in our architecture (shown in Figure \ref{fig:architecture}).
\subsection{Text preprocessing}
We adopt preprocessing steps tailored to the characteristics of ASR transcriptions. 
Consecutive repeated unigrams and bigrams are reduced to single terms.
Specific ASR tags, such as \textit{\{vocalsound\}, \{pause\}, and \{gap\}} are filtered out.
In addition, filler words, such as \textit{uh-huh}, \textit{okay}, \textit{well}, and \textit{by the way} are also discarded.
Consecutive stopwords at the beginning and end of utterances are stripped.
In the end, utterances that contain less than 3 non-stopwords are pruned out. The surviving utterances are used for the next steps.

\subsection{Utterance community detection} \label{subsec:utterance_community_detection}

The goal here is to cluster utterances into communities that should be summarized by a common abstractive sentence. 

We initially experimented with techniques capitalizing on word vectors, such as $k$-means and hierarchical clustering based on the Euclidean distance or the Word Mover's Distance \cite{kusner2015word}. We also tried graph-based approaches, such as community detection in a complete graph where nodes are utterances and edges are weighted based on the aforementioned distances.

Best results were obtained, however, with a simple approach in which utterances are projected into the vector space and assigned standard TF-IDF weights. Then, the dimensionality of the utterance-term matrix is reduced with Latent Semantic Analysis (LSA), and finally, the $k$-means algorithm is applied. Note that LSA is only used here, during the utterance community detection phase, to remove noise and stabilize clustering. We do not use a topic graph in our approach.

We think using word embeddings was not effective, because in meeting speech, as opposed to traditional documents, participants tend to use the same term to refer to the same thing throughout the entire conversation, as noted by Riedhammer et al. \shortcite{riedhammer2010long}, and as verified in practice. This is probably why, for clustering utterances, capturing synonymy is counterproductive, as it artificially reduces the distance between every pair of utterances and blurs the picture.
\subsection{Multi-Sentence Compression} \label{subsec:msc}
The following steps are performed separately for each community.

\subsubsection*{Word importance scoring}\label{subsub:deg}
From a processed version of the community (stemming and stopword removal), we construct an undirected, weighted word co-occurrence network as described in subsection \ref{app:a}. We use a sliding window of size $W=6$ not overspanning utterances. Note that stemming is performed only here, and for the sole purpose of building the word co-occurrence network.

We then compute the CoreRank numbers of the nodes as described in subsection \ref{app:a}.

We finally reweigh the CoreRank scores, indicative of word importance within a given community, with a quantity akin to an \textit{Inverse Document Frequency}, where communities serve as documents and the full meeting as the collection. We thus obtain something equivalent to the TW-IDF weighting scheme of Rousseau and Vazirgiannis~\shortcite{rousseau2013graph}, where the CoreRank scores are the term weights TW:

\vspace{-0.75cm}

\begin{equation} \label{eq:tw-idf}
TW\text{-}IDF(t, d, D)=TW(t, d) \times IDF(t, D)
\end{equation}

where $t$ is a term belonging to community $d$, and $D$ is the set of all utterance communities. We compute the IDF as $IDF(t,D) = 1 + \mathrm{log}\nicefrac{|D|}{D_t}$, where $|D|$ is the number of communities and  $D_t$ the number of communities containing $t$.

The intuition behind this reweighing scheme is that a term should be considered important within a given meeting if it has a high CoreRank score within its community \textit{and} if the number of communities in which the term appears is relatively small.

\subsubsection*{Word graph building}
The backbone of the graph is laid out as a directed sequence of nodes corresponding to the words in the first utterance, with special \texttt{START} and \texttt{END} nodes at the beginning and at the end (see Figure \ref{fig:msc_illustration}). Edge direction follows the natural flow of text. Words from the remaining utterances are then iteratively added to the graph (between the \texttt{START} and \texttt{END} nodes) based on the following rules:\\

\vspace{-0.2cm}

1) if the word is a \textbf{non-stopword}, the word is mapped onto an existing node if it has the same lowercased form and the same part-of-speech tag\footnote{\tiny{We used NLTK's averaged perceptron tagger, available at: \url{http://www.nltk.org/api/nltk.tag.html\#module-nltk.tag.perceptron}}}. In case of multiple matches, we check the immediate context (the preceding and following words in the utterance and the neighboring nodes in the graph), and we pick the node with the largest context overlap or which has the greatest number of words already mapped to it (when no overlap). When there is no match, we use WordNet as described in Appendix \ref{app:word_net}.

2) if the word is a \textbf{stopword} and there is a match, it is mapped only if there is an overlap of at least one non-stopword in the immediate context. Otherwise, a new node is created.\\

\vspace{-0.35cm}

\noindent Finally, note that any two words appearing within the same utterance cannot be mapped to the same node. This ensures that every utterance is a loopless path in the graph. Of course, there are many more paths in the graphs than original utterances.

\begin{figure*}[!ht]
\centering
\captionsetup{size=small}
\begin{minipage}[c]{0.38\textwidth}
\caption{
t-SNE visualization \cite{maaten2008visualizing} of the Google News vectors of the words in the utterance community shown in Figure \ref{fig:msc_illustration}. Arrows join the words in the best compression path shown in Figure \ref{fig:msc_illustration}. Movements in the embedding space, as measured by the number of unique clusters covered by the path (here, $6/11$), provide a sense of the diversity of the compressed sentence, as formalized in Equation \ref{eq:div}.
} \label{fig:div_illustration}
\end{minipage}\hfill
\begin{minipage}[c]{0.6\textwidth}
\includegraphics[scale=0.385]{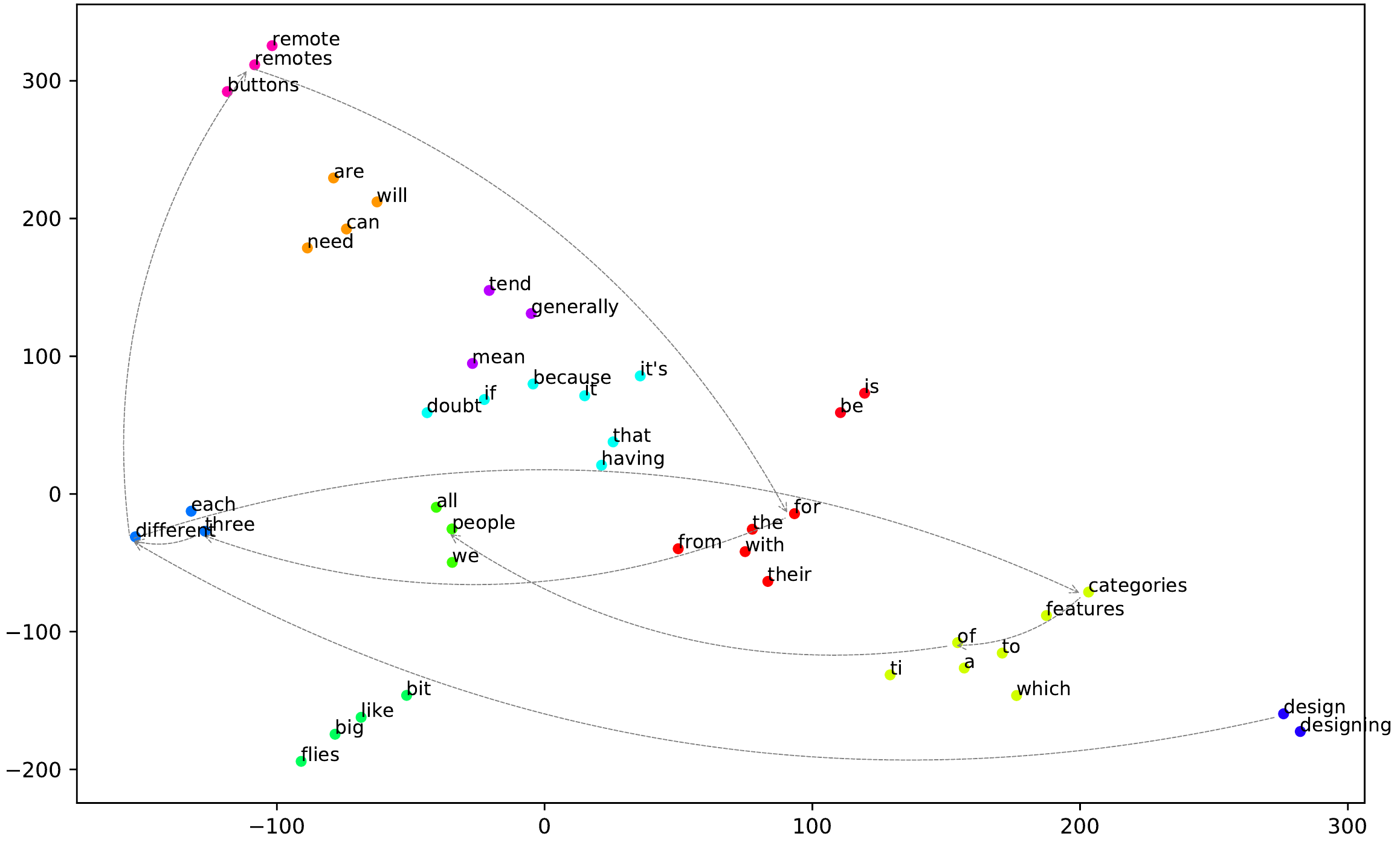}
\end{minipage}

\end{figure*}

\subsubsection*{Edge Weight Assignment}
Once the word graph is constructed, we assign weights to its edges as: 

\begin{equation}
w^{\prime\prime\prime} (p_i, p_j) = \frac{w^\prime (p_i, p_j)}{w^{\prime\prime} (p_i, p_j)}
\end{equation}\label{eq:e_w}

\noindent where $p_i$ and $p_j$ are two neighbors in the MSCG. As detailed next, those weights combine \textit{local co-occurrence statistics} (numerator) with \textit{global exterior knowledge} (denominator). Note that the lower the weight of an edge, the better.

\textbf{\textit{Local co-occurrence statistics}}. \\
We use Filippova \shortcite{filippova2010multi}'s formula:
\begin{equation}
w^\prime (p_i, p_j) = \frac{f(p_i) + f(p_j)}{\sum_{P \in G', p_i, p_j \in P} \mathrm{diff}(P, p_i, p_j)^{- 1}}
\end{equation}
where $f(p_i)$ is the number of words mapped to node $p_i$ in the MSCG $G'$, and $\mathrm{diff}(P, p_i, p_j)^{- 1}$ is the inverse of the distance between $p_i$ and $p_j$ in a path $P$ (in number of hops). This weighting function favors edges between infrequent words that frequently appear close to each other in the text (the lower, the better).

\textbf{\textit{Global exterior knowledge}}. \\
We introduce a second term based on the \textit{Word Attraction Force score} of Wang et al. \shortcite{wang2014corpus}:
\begin{equation}
w^{\prime\prime} (p_i, p_j) = \frac{f(p_i) \times f(p_j)}{d_{p_i,p_j}^2}
\end{equation}
where $d_{p_i,p_j}$ is the Euclidean distance between the words mapped to $p_i$ and $p_j$ in a word embedding space\footnote{\tiny{GoogleNews vectors \url{https://code.google.com/archive/p/word2vec}}}. This component favor paths going through salient words that have \textit{high semantic similarity} (the higher, the better). The goal is to ensure readability of the compression, by avoiding to generate a sentence jumping from one word to a completely unrelated one.

\subsubsection*{Path re-ranking}
As in Boudin and Morin \shortcite{boudin2013keyphrase}, we use a shortest weighted path algorithm to find the $K$ paths between the \texttt{START} and \texttt{END} symbols having the lowest cumulative edge weight:
\begin{equation}
W (P) = \sum_{i = 1}^{|P| - 1} w^{\prime\prime\prime} (p_i, p_{i + 1})
\end{equation}
Where $|P|$ is the number of nodes in the path. Paths having less than $z$ words or that do not contain a verb are filtered out ($z$ is a tuning parameter). However, unlike in Boudin and Morin \shortcite{boudin2013keyphrase}, we rerank the $K$ best paths with the following novel weighting scheme (the lower, the better), and the path with the lowest score is used as the compression:

\begin{equation}
\begin{aligned}
\mathrm{score}(P) = \frac{W(P)}{|P| \times F(P) \times C(P) \times D(P)}
\end{aligned}
\end{equation}

\noindent The denominator takes into account the length of the path, and its fluency ($F$), coverage ($C$), and diversity ($D$). $F$, $C$, and $D$ are detailed in what follows.

\textbf{\textit{Fluency}}. We estimate the grammaticality of a path with an $n$-gram language model. In our experiments, we used a trigram model\footnote{\tiny{CMUSphinx English LM: \url{https://cmusphinx.github.io}}}:

\begin{equation}
\begin{aligned}
F(P) = \frac{\sum^{|P|}_{i = 1} \mathrm{log} Pr (p_i |p_{i - n + 1}^{i - 1})}{\#n\text{-}gram}
\end{aligned}
\end{equation}
where $|P|$ denote path length, and $p_i$ and $\#n\text{-}gram$ are respectively the words and number of $n$-grams in the path.

\textbf{\textit{Coverage}}. We reward the paths that visit important nouns, verbs and adjectives:
\begin{equation} \label{eq:7}
C(P) = \frac{\sum_{p_i \in P} \mathrm{TW\text{-}IDF} (p_i)}{\#p_i}
\end{equation}
where $\#p_i$ is the number of nouns, verbs and adjectives in the path. The TW-IDF scores are computed as explained in subsection \ref{subsub:deg}.

\textbf{\textit{Diversity}}. We cluster all words from the MSCG in the word embedding space by applying the $k$-means algorithm. We then measure the diversity of the vocabulary contained in a path as the number of unique clusters visited by the path, normalized by the length of the path:

\begin{equation} \label{eq:div}
D(P) = \frac{\sum^k_{j = 1} 1_{\exists p_i \in P | p_i \in \mathrm{cluster}_j}}{|P|}
\end{equation}
The graphical intuition for this measure is provided in Figure \ref{fig:div_illustration}. Note that we do not normalize $D$ by the total number of clusters (only by path length) because $k$ is fixed for all candidate paths.

\subsection{Budgeted submodular maximization} \label{subsec:submodularity}
We apply the previous steps separately for all utterance communities, which results in a set $\mathcal{S}$ of abstractive sentences (one for each community). This set of sentences can already be considered to be a summary of the meeting. However, it might exceed the maximum size allowed, and still contain some redundancy or off-topic sections unrelated to the general theme of the meeting (e.g., chit-chat). 

Therefore, we design the following \textit{submodular} and \textit{monotone non-decreasing} objective function:
\begin{equation}
\begin{aligned}
f(S) = \sum_{s_i \in S} n_{s_i} w_{s_i} + \lambda \sum^k_{j = 1} 1_{\exists s_i \in S | s_i \in cluster_j}
\end{aligned}
\label{eq:obj}
\end{equation}
where $\lambda \geq 0$ is the trade-off parameter, $n_{s_i}$ is the number of occurrences of word $s_i$ in $S$, and $w_{s_i}$ is the CoreRank score of $s_i$.

Then, as explained in subsection \ref{sub:rel_sub}, we obtain a near-optimal subset of abstractive sentences by maximizing $f$ with a greedy algorithm. CoreRank scores and clusters are found as previously described, except that this time they are obtained from the full processed meeting transcription rather than from a single  utterance community.

\section{Experimental setup}

\subsection{Datasets} \label{subsec:datasets}
We conducted experiments on the widely-used AMI \cite{mccowan2005ami} and ICSI \cite{janin2003icsi} benchmark datasets. We used the traditional test sets of 20 and 6 meetings respectively for the AMI and ICSI corpora \cite{riedhammer2008packing}. Each meeting in the AMI test set is associated with a human abstractive summary of 290 words on average, whereas each meeting in the ICSI test set is associated with 3 human abstractive summaries of respective average sizes 220, 220 and 670 words. 

For parameter tuning, we constructed development sets of 47 and 25 meetings, respectively for AMI and ICSI, by randomly sampling from the training sets. The word error rate of the ASR transcriptions is respectively of 36\% and 37\% for AMI and ICSI.

\subsection{Baselines}
We compared our system against 7 baselines, which are listed below and more thoroughly detailed in Appendix \ref{app:baselines}. Note that preprocessing was exactly the same for our system and all baselines. \\
$\bullet$ \textbf{Random} and \textbf{Longest Greedy} are basic baselines recommended by  \cite{riedhammer2008packing}, \\ 
$\bullet$ \textbf{TextRank} \cite{mihalcea2004textrank}, \\
$\bullet$ \textbf{ClusterRank} \cite{garg2009clusterrank}, \\
$\bullet$ \textbf{CoreRank \& PageRank submodular} \cite{tixier2017combining}, \\
$\bullet$ \textbf{Oracle} is the same as the random baseline, but uses the human extractive summaries as input.

In addition to the baselines above, we included in our comparison 3 variants of our system using different MSCGs: \textbf{Our System (Baseline)} uses the original MSCG of Filippova \shortcite{filippova2010multi}, \textbf{Our System (KeyRank)} uses that of Boudin and Morin~\shortcite{boudin2013keyphrase}, and \textbf{Our System (FluCovRank)} that of Mehdad et al. \shortcite{mehdad2013abstractive}. Details about each approach were given in Section \ref{sec:related_work}.

\subsection{Parameter tuning}
For \textit{Our System} and each of its variants, we conducted a grid search on the development sets of each corpus, for fixed summary sizes of 350 and 450 words (AMI and ICSI). We searched the following parameters:

\noindent $\bullet$ $n$: number of utterance communities (see Section \ref{subsec:utterance_community_detection}). We tested values of $n$ ranging from 20 to 60, with steps of 5. This parameter controls how much abstractive should the summary be. If all utterances are assigned to their own singleton community, the MSCG is of no utility, and our framework is extractive. It becomes more and more abstractive as the number of communities decreases. \\
$\bullet$ $z$: minimum path length (see Section \ref{subsec:msc}). We searched values in the range $[6,16]$ with steps of 2. If a path is shorter than a certain minimum number of words, it often corresponds to an invalid sentence, and should thereby be filtered out. \\
$\bullet$ $\lambda$ and $r$, the trade-off parameter and the scaling factor (see Section \ref{subsec:submodularity}). We searched $[0, 1]$ and $[0, 2]$ (respectively) with steps of 0.1. The parameter $\lambda$ plays a regularization role favoring diversity. The scaling factor makes sure the quality function gain and utterance cost are comparable.

The best parameter values for each corpus are summarized in Table \ref{table:optimal_parameters}. $\lambda$ is mostly non-zero, indicating  that it is necessary to include a regularization term in the submodular function. In some cases though, $r$ is equal to zero, which means that utterance costs are not involved in the greedy decision heuristic. These observations contradict the conclusion of Lin \shortcite{lin2012submodularity} that $r=0$ cannot give best results.

\begin{table}[!ht]
\setlength{\tabcolsep}{3.2pt}
\small
\centering
\scalebox{0.95}{
\begin{tabular}{rrr}
\hline
	System & AMI & ICSI \\ 
\hline
    Our System                       & 50, 8, (0.7, 0.5) & 40, 14, (0.0, 0.0) \\
    Our System (Baseline)            & 50, 12, (0.3, 0.5) & 45, 14, (0.1, 0.0) \\
    Our System (KeyRank)             & 50, 10, (0.2, 0.9) & 45, 12, (0.3, 0.4) \\
    Our System (FluCovRank)          & 35, 6, (0.4, 1.0) & 50, 10, (0.2, 0.3) \\
\hline
\end{tabular}
}
\caption{Optimal parameter values $n, z, (\lambda, r)$.}
\label{table:optimal_parameters}
\end{table}

Apart from the tuning parameters, we set the number of LSA dimensions to 30 and 60 (resp. on AMI and ISCI). The small number of LSA dimensions retained can be explained by the fact that the AMI and ICSI transcriptions feature 532 and 1126 unique words on average, which is much smaller than traditional documents. This is due to relatively small meeting duration, and to the fact that participants tend to stick to the same terms throughout the entire conversation. For the $k$-means algorithm, $k$ was set equal to the minimum path length $z$ when doing MSCG path re-ranking (see Equation \ref{eq:div}), and to 60 when generating the final summary (see Equation \ref{eq:obj}).

Following Boudin and Morin \shortcite{boudin2013keyphrase}, the number of shortest weighted paths $K$ was set to 200, which is greater than the $K=100$ used by Filippova \shortcite{filippova2010multi}. Increasing $K$ from 100 improves performance with diminishing returns, but significantly increases complexity. We empirically found 200 to be a good trade-off.

\section{Results and Interpretation}

\begin{figure*}[ht]
\centering
\begin{subfigure}[t]{0.49\textwidth}
\centering
\includegraphics[scale=0.31]{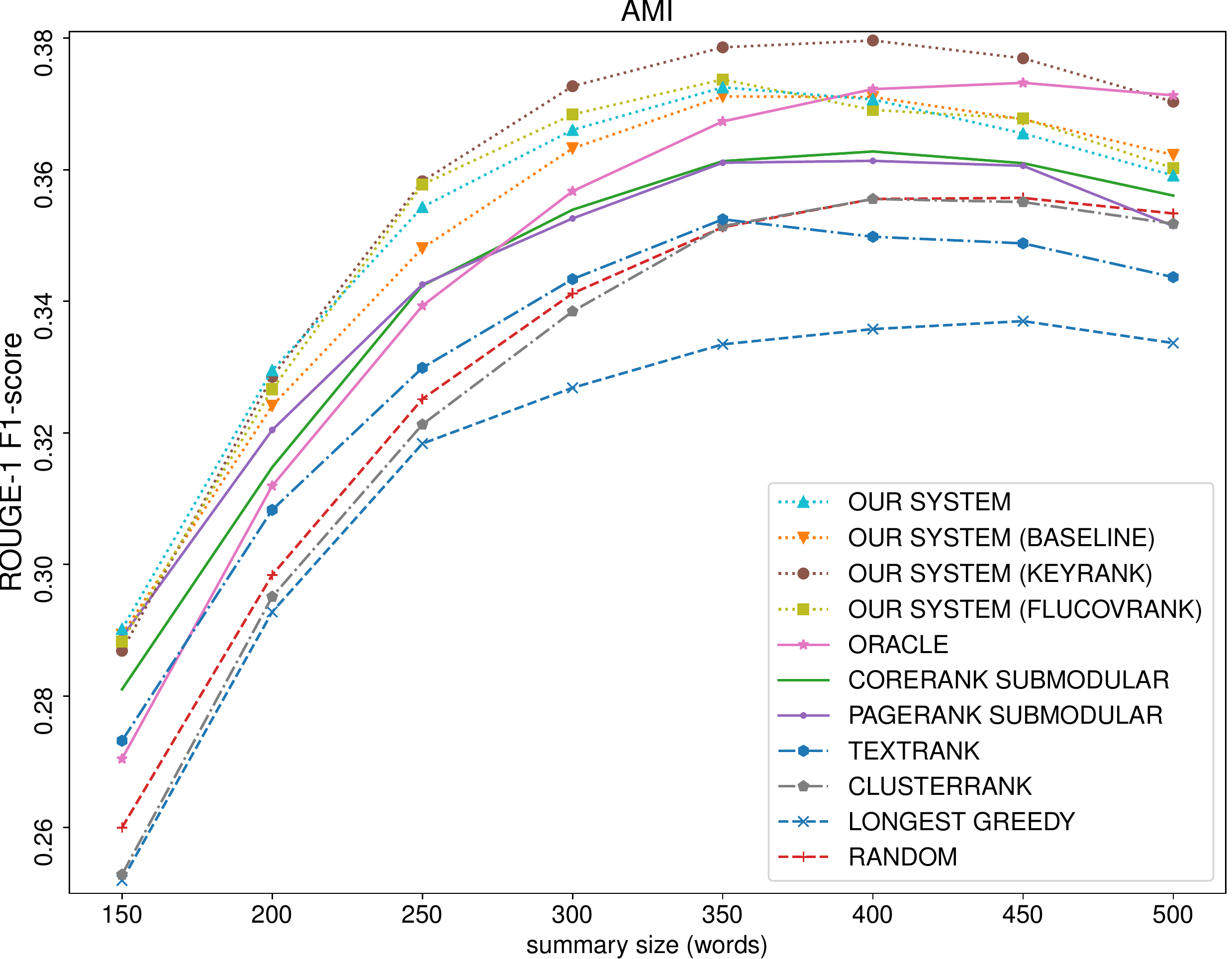}
\end{subfigure}
\begin{subfigure}[t]{0.49\textwidth}
\centering
\includegraphics[scale=0.31]{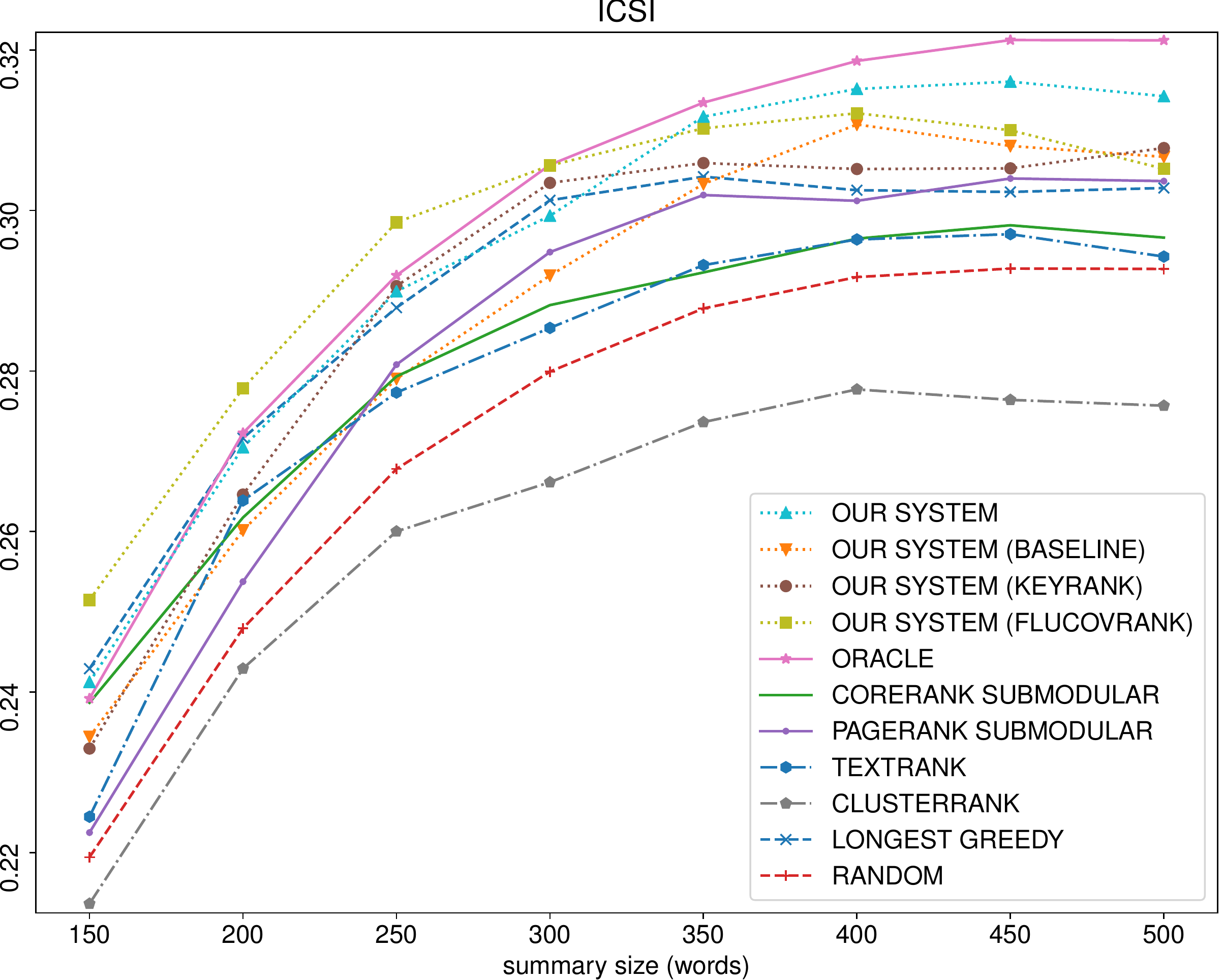}
\end{subfigure}
\caption{ROUGE-1 F-1 scores for various budgets (ASR transcriptions).}
\label{fig:res}
\end{figure*}

\noindent\textbf{Metrics}.
We evaluated performance with the widely-used ROUGE-1, ROUGE-2 and ROUGE-SU4 metrics \cite{lin2004rouge}. These metrics are respectively based on unigram, bigram, and unigram plus skip-bigram overlap with maximum skip distance of 4, and have been shown to be highly correlated with human evaluations \cite{lin2004rouge}. ROUGE-2 scores can be seen as a measure of summary readability \cite{lin2003automatic,ganesan2010opinosis}. ROUGE-SU4 does not require consecutive matches but is still sensitive to word order.

Macro-averaged results for summaries generated from automatic transcriptions can be seen in Figure \ref{fig:res} and Table \ref{table:res}. Table \ref{table:res} provides detailed comparisons over the fixed budgets that we used for parameter tuning, while Figure \ref{fig:res} shows the performance of the models for budgets ranging from 150 to 500 words. The same information for summaries generated from manual transcriptions is available in Appendix \ref{app:res_manual}. Finally, summary examples are available in Appendix \ref{app:examples}.

\begin{table*}[ht]
\small
\centering
\setlength{\tabcolsep}{2pt} 
\renewcommand{\arraystretch}{1.1} 
\scalebox{0.87}{
\begin{tabular}{|r||ccc|ccc|ccc||ccc|ccc|ccc|}
  \hline
  \multirow{3}{*}{}
      & \multicolumn{3}{c|}{AMI ROUGE-1} 
      & \multicolumn{3}{c|}{AMI ROUGE-2}
      & \multicolumn{3}{c||}{AMI ROUGE-SU4} 
      & \multicolumn{3}{c|}{ICSI ROUGE-1} 
      & \multicolumn{3}{c|}{ICSI ROUGE-2} 
      & \multicolumn{3}{c|}{ICSI ROUGE-SU4}
      \\  \cline{2-19}         
                                    & R & P & F-1 & R & P & F-1 & R & P & F-1 & R & P & F-1 & R & P & F-1 & R & P & F-1 \\  \hline
    Our System                      & 41.83 & 34.44 & 37.25 & 8.22 & 6.95 & 7.43 & 15.83 & 13.70 & 14.51 & 36.99 & 28.12 & \textbf{31.60} & 5.41 & 4.39 & 4.79 & 13.10 & 10.17 & \textbf{11.35} \\
    Our System (Baseline)           & 41.56 & 34.37 & 37.11 & 7.88 & 6.66 & 7.11 & 15.36 & 13.20 & 14.02 & 36.39 & 27.20 & 30.80 & 5.19 & 4.12 & 4.55 & 12.59 & 9.70 & 10.86 \\
    Our System (KeyRank)            & 42.43 & 35.01 & \textbf{37.86} & 8.72 & 7.29 & \textbf{7.84} & 16.19 & 13.76 & \textbf{14.71} & 35.95 & 27.00 & 30.52 & 4.64 & 3.64 & 4.04 & 12.43 & 9.23 & 10.50 \\
    Our System (FluCovRank)         & 41.84 & 34.61 & 37.37 & 8.29 & 6.92 & 7.45 & 16.28 & 13.48 & 14.58 & 36.27 & 27.56 & 31.00 & 5.56 & 4.35 & \textbf{4.83} & 13.47 & 9.85 & 11.29 \\
\hline
    Oracle                          & 40.49 & 34.65 & \textbf{36.73} & 8.07 & 7.35 & \textbf{7.55} & 15.00 & 14.03 & \textbf{14.26} & 37.91 & 28.39 & \textbf{32.12} & 5.73 & 4.82 & \textbf{5.18} & 13.35 & 10.73 & \textbf{11.80} \\
    CoreRank Submodular             & 41.14 & 32.93 & 36.13 & 8.06 & 6.88 & 7.33 & 14.84 & 13.91 & 14.18 & 35.22 & 26.34 & 29.82 & 4.36 & 3.76 & 4.00 & 12.11 & 9.58 & 10.61 \\
    PageRank Submodular             & 40.84 & 33.08 & 36.10 & 8.27 & 6.88 & 7.42 & 15.37 & 13.71 & 14.32 & 36.05 & 26.69 & 30.40 & 4.82 & 4.16 & 4.42 & 12.19 & 10.39 & 11.14 \\
    TextRank                        & 39.55 & 32.60 & 35.25 & 7.67 & 6.43 & 6.90 & 14.87 & 12.87 & 13.62 & 34.89 & 26.33 & 29.70 & 4.60 & 3.74 & 4.09 & 12.42 & 9.43 & 10.64 \\
    ClusterRank                     & 39.36 & 32.53 & 35.14 & 7.14 & 6.05 & 6.46 & 14.34 & 12.80 & 13.35 & 32.63 & 24.44 & 27.64 & 4.03 & 3.44 & 3.68 & 11.04 & 8.88 & 9.77 \\
    Longest Greedy                  & 37.31 & 30.93 & 33.35 & 5.77 & 4.71 & 5.11 & 13.79 & 11.11 & 12.15 & 35.57 & 26.74 & 30.23 & 4.84 & 3.88 & 4.27 & 13.09 & 9.46 & 10.90 \\
    Random                          & 39.42 & 32.48 & 35.13 & 6.88 & 5.89 & 6.26 & 14.07 & 12.70 & 13.17 & 34.78 & 25.75 & 29.28 & 4.19 & 3.51 & 3.78 & 11.61 & 9.37 & 10.29 \\
\hline
\end{tabular}
}
\caption{Macro-averaged results for 350 and 450 word summaries (ASR transcriptions). \label{table:res}}
\end{table*}

\noindent \textbf{ROUGE-1}.
Our systems outperform all baselines on AMI (including \textit{Oracle}) and all baselines on ICSI (except \textit{Oracle}). Specifically, \textit{Our System} is best on ICSI, while \textit{Our System (KeyRank)} is superior on AMI. We can also observe on Figure \ref{fig:res} that our systems are consistently better throughout the different summary sizes, even though their parameters were tuned for specific sizes only. This shows that the best parameter values are quite robust across the entire budget range.

\noindent \textbf{ROUGE-2}. Again, our systems (except \textit{Our System (Baseline)}) outperform all baselines, except \textit{Oracle}. In addition, \textit{Our System} and \textit{Our~System~(FluCovRank)} consistently improve on \textit{Our System (Baseline)}, which proves that the novel components we introduce improve summary fluency. 

\noindent \textbf{ROUGE-SU4}.
ROUGE-SU4 was used to measure the amount of in-order word pairs overlapping. Our systems are competitive with all baselines, including \textit{Oracle}. Like with ROUGE-1, \textit{Our System} is better than \textit{Our System (KeyRank)} on ICSI, whereas the opposite is true on AMI. 

\noindent \textbf{General remarks}.\\
\noindent $\bullet$ The summaries of all systems except \textit{Oracle} were generated from noisy ASR transcriptions, but were compared against human abstractive summaries. ROUGE being based on word overlap, it makes it very difficult to reach very high scores, because many words in the ground truth summaries do not appear in the transcriptions at all.

\noindent $\bullet$ The scores of all systems are lower on ICSI than on AMI. This can be explained by the fact that on ICSI, the system summaries have to jointly match 3 human abstractive summaries of different content and size, which is much more difficult than matching a single summary.

\noindent $\bullet$ Our framework is very competitive to \textit{Oracle}, which is notable since the latter has direct access to the human extractive summaries. Note that \textit{Oracle} does not reach very high ROUGE scores because the overlap between the human extractive and abstractive summaries is low (19\% and 29\%, respectively on AMI and ICSI test sets).

\section{Conclusion and Next Steps}
\vspace{-0.1cm}
Our framework combines the strengths of 6 approaches that had previously been applied to 3 different tasks (keyword extraction, multi-sentence compression, and summarization) into a unified, fully unsupervised end-to-end summarization framework, and introduces some novel components. Rigorous evaluation on the AMI and ICSI corpora shows that we reach state-of-the-art performance, and generate reasonably grammatical abstractive summaries despite taking noisy utterances as input and not relying on any annotations or training data. Finally, thanks to its fully unsupervised nature, our method is applicable to other languages than English in an almost out-of-the-box manner.

Our framework was developed for the meeting domain. Indeed, our generative component, the multi-sentence compression graph (MSCG), needs redundancy to perform well. Such redundancy is typically present in meeting speech but not in traditional documents. In addition, the MSCG is by design robust to noise, and our custom path re-ranking strategy, based on graph degeneracy, makes it even more robust to noise. As a result, our framework is advantaged on ASR input. Finally, we use a language model to favor fluent paths, which is crucial when working with (meeting) speech but not that important when dealing with well-formed input.

Future efforts should be dedicated to improving the community detection phase and generating more abstractive sentences, probably by harnessing Deep Learning. However, the lack of large training sets for the meeting domain is an obstacle to the use of neural approaches.

\vspace{-0.15cm}

\section*{Acknowledgments}
We are grateful to the three anonymous reviewers for their detailed and constructive feedback. This research was supported in part by the OpenPaaS::NG project. 

\bibliography{main}
\bibliographystyle{acl_natbib}

\onecolumn
\newpage
\begin{center}
\textbf{\Large Supplementary Material}
\end{center}
\textbf{\Large Appendices}
\appendix
\section{Use of WordNet} \label{app:word_net}

When the word to be mapped to the MSCG is a \textbf{non-stopword}, and if there is no node in the graph that has the same lowercased form and the same part-of-speech tag, we try to perform the mapping by using WordNet in the following order:

\begin{enumerate}[label=(\roman*), leftmargin=*]
\setlength\itemsep{0em}
\item there is a node which is a synonym of the word (e.g., ``price'' and ``costs''). The word is mapped to that node, and the node is relabeled with the word if the latter has a higher TW-IDF score.

\item there is a node which is a hypernym of the word (e.g., ``diamond'' and ``gemstone''). The word is mapped to that node, and the node is relabeled with the word if the latter has a higher TW-IDF score.

\item there is a node which shares a common hypernym with the word (e.g., ``red'',``blue'' $\rightarrow$ ``color''). If the product of the WordNet path distance similarities of the common hypernym with the node and the word exceeds a certain threshold, the word is mapped to that node and the node is relabeled with the hypernym. A completely new word might thus be introduced. We set its TW-IDF score as the highest TW-IDF of the two words it replaces. When multiple nodes are eligible for mapping, we select the one with greatest path distance similarity product.

\item there is a node which is in an entailment relation with the word (e.g., ``look'' is entailed by ``see''). The word is mapped to that node, and the node is relabeled with the word if the latter has a higher TW-IDF score.
\end{enumerate}

\noindent In attempts \textbf{i}, \textbf{ii}, and \textbf{iv} above, if there is more than one candidate node, we select the one with highest TW-IDF score. If all attempts above are unsuccessful, a new node is created for the word.

\section{Baseline Details}\label{app:baselines}

\begin{itemize}[leftmargin=*]
\setlength\itemsep{0em}
\item \textbf{Random}. A basic baseline recommended by \cite{riedhammer2008packing} to ease cross-study comparison. This system randomly selects utterances without replacement from the transcription until the budget is violated. To account for stochasticity, we report scores averaged over 30 runs.

\item \textbf{Longest Greedy}. A basic baseline recommended by \cite{riedhammer2008packing} to ease cross-study comparison. The longest remaining utterance is selected at each step from the transcription until the summary size constraint is satisfied.

\item \textbf{TextRank} \cite{mihalcea2004textrank}. Utterances within the transcription are represented as nodes in an undirected complete graph, and edge weights are assigned based on lexical similarity between utterances. To provide a summary, the top nodes according to the weighted PageRank algorithm \cite{page1999pagerank} are selected. We used a publicly available implementation\footnote{\url{https://github.com/summanlp/textrank}}.

\item \textbf{ClusterRank} \cite{garg2009clusterrank}. This system is an extension of TextRank to meeting summarization. Firstly, utterances are segmented into clusters. A complete graph is built from the clusters. Then, a score is assigned to each utterance based on both the PageRank score of the cluster it belongs to and its cosine similarity with the cluster centroid. In the end, a greedy selection strategy is applied to build the summary out of the highest scoring utterances. Since the authors did not make their code publicly available and were not able to share it privately, we wrote our own implementation.

\item \textbf{CoreRank submodular \& PageRank submodular} \cite{tixier2017combining}. These two \textit{extractive} baselines implement the last step of our pipeline (see Section \ref{subsec:submodularity}). That is, budgeted submodular maximization is applied directly on the full list of utterances. As can be inferred from their names, the only difference between those two baselines is that the first uses PageRank scores, whereas the second uses CoreRank scores.

\item \textbf{Oracle}. This system is the same as the Random baseline, but instead of sampling utterances from the ASR transcription, it draws from the human extractive summaries. Annotators put those summaries together by selecting the best utterances from the entire manual transcription. Scores were averaged over 30 runs due to the randomness of the procedure.
\end{itemize}

\section{Results for Manual Transcriptions}\label{app:res_manual}

\begin{figure*}[ht]
\centering

\begin{subfigure}[t]{0.49\textwidth}
\centering
\includegraphics[scale=0.31]{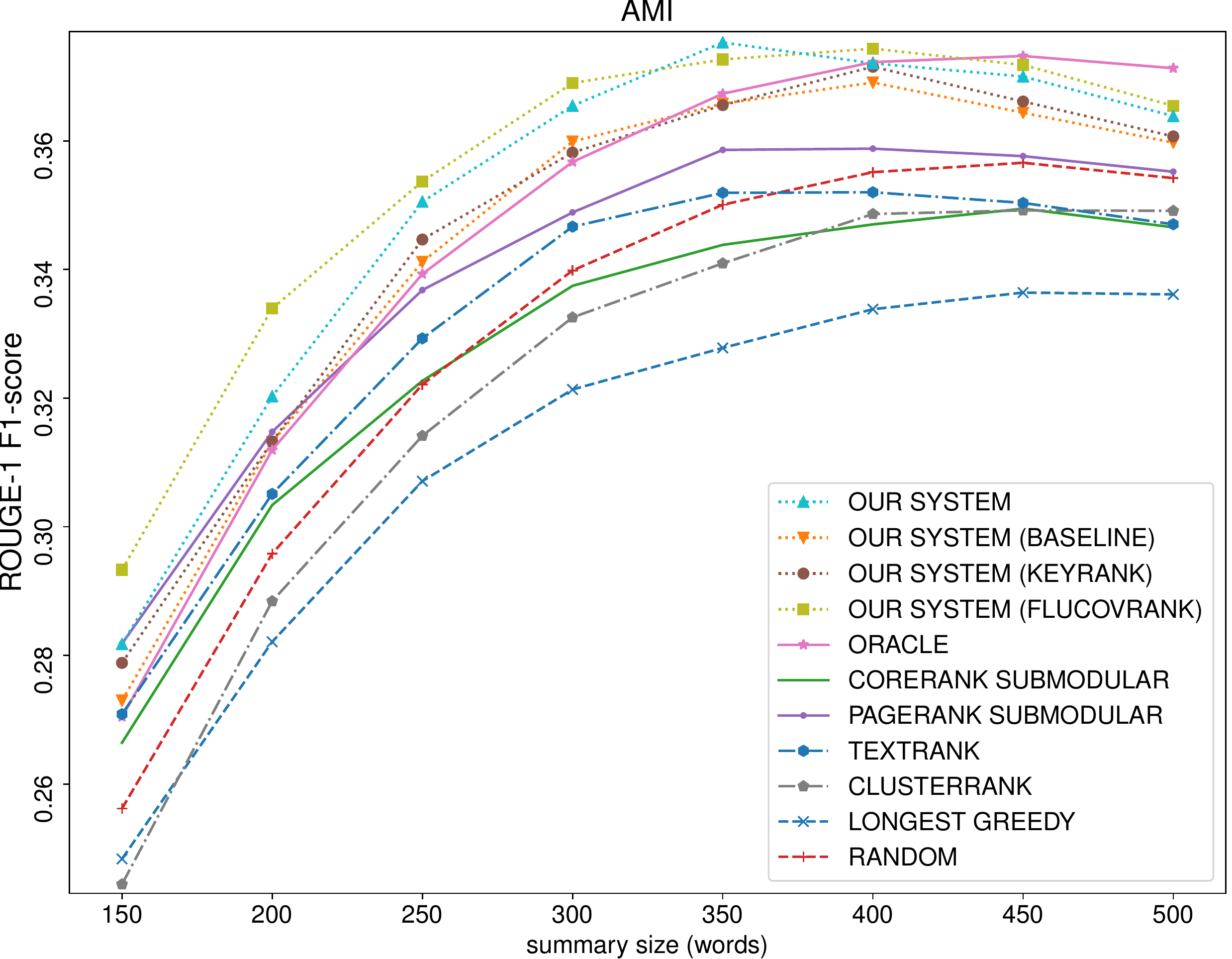}
\end{subfigure}
\begin{subfigure}[t]{0.49\textwidth}
\centering
\includegraphics[scale=0.31]{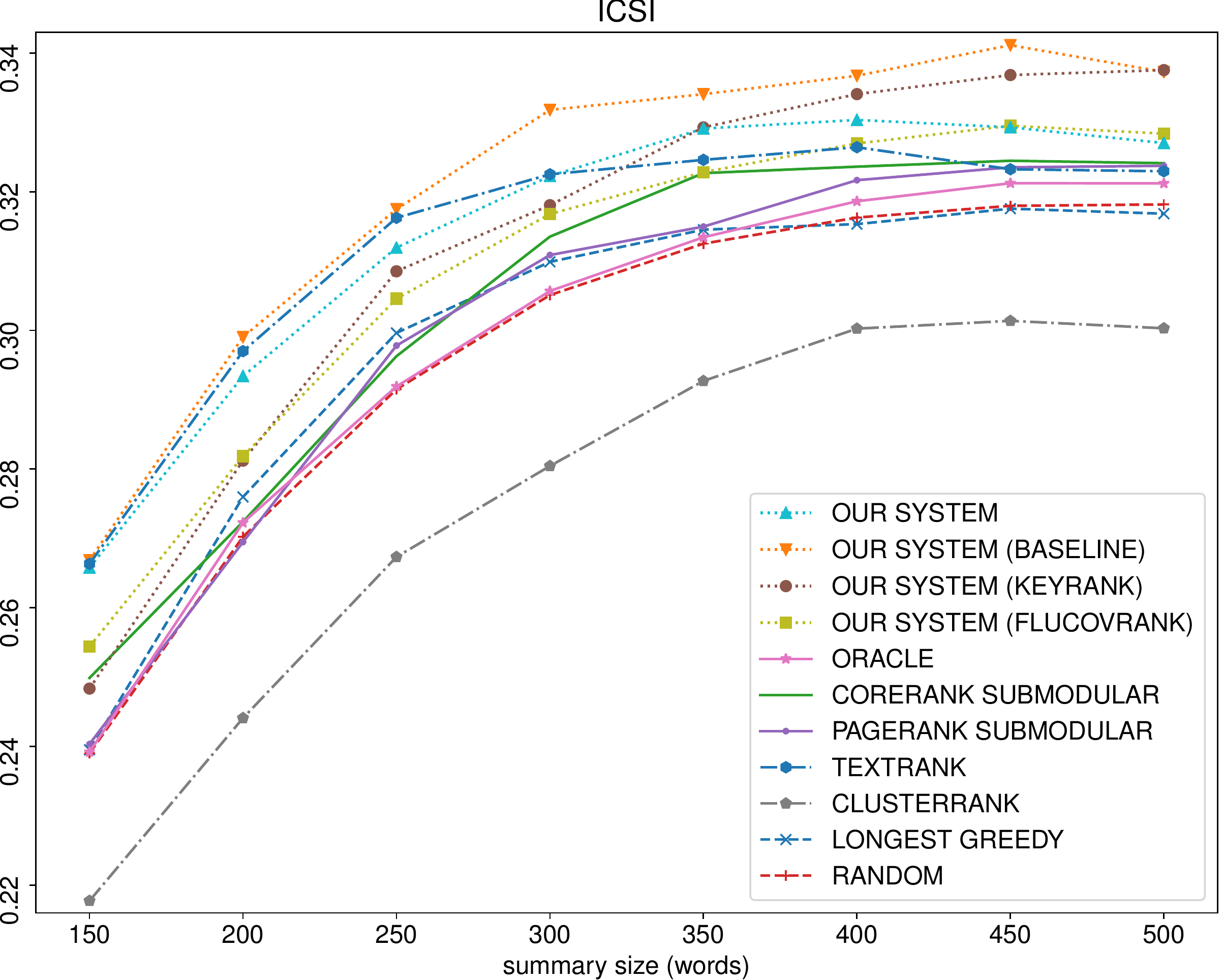}
\end{subfigure}
\caption{ROUGE-1 F-1 scores for various budgets (manual transcriptions).}
\label{fig:res_manual}
\end{figure*}

\begin{table*}[!ht]
\small
\centering
\setlength{\tabcolsep}{2pt} 
\renewcommand{\arraystretch}{1.1} 
\scalebox{0.87}{
\begin{tabular}{|r||ccc|ccc|ccc||ccc|ccc|ccc|}
  \hline
  \multirow{3}{*}{}
      & \multicolumn{3}{c|}{AMI ROUGE-1} 
      & \multicolumn{3}{c|}{AMI ROUGE-2}
      & \multicolumn{3}{c||}{AMI ROUGE-SU4} 
      & \multicolumn{3}{c|}{ICSI ROUGE-1} 
      & \multicolumn{3}{c|}{ICSI ROUGE-2} 
      & \multicolumn{3}{c|}{ICSI ROUGE-SU4}
      \\  \cline{2-19}         
                                    & R & P & F-1 & R & P & F-1 & R & P & F-1 & R & P & F-1 & R & P & F-1 & R & P & F-1 \\  \hline
    Our System                      & 42.03 & 34.77 & \textbf{37.53} & 8.87 & 7.56 & 8.06 & 15.92 & 14.08 & \textbf{14.76} & 38.57 & 29.30 & 32.93 & 5.80 & 4.74 & 5.14 & 13.92 & 10.79 & 12.04 \\
    Our System (Baseline)           & 40.88 & 33.96 & 36.58 & 8.13 & 6.95 & 7.39 & 15.17 & 13.25 & 13.97 & 40.03 & 30.20 & \textbf{34.11} & 6.65 & 5.51 & \textbf{5.98} & 14.65 & 11.37 & \textbf{12.70} \\
    Our System (KeyRank)            & 40.87 & 33.91 & 36.56 & 8.42 & 7.12 & 7.62 & 15.50 & 13.48 & 14.25 & 39.55 & 29.79 & 33.68 & 6.32 & 5.19 & 5.64 & 14.63 & 10.99 & 12.47 \\
    Our System (FluCovRank)         & 41.73 & 34.50 & 37.27 & 8.45 & 7.05 & 7.60 & 16.08 & 13.47 & 14.49 & 38.57 & 29.21 & 32.95 & 6.38 & 5.08 & 5.60 & 14.38 & 10.62 & 12.13 \\
\hline
    Oracle                          & 40.49 & 34.65 & \textbf{36.73} & 8.07 & 7.35 & \textbf{7.55} & 15.00 & 14.03 & \textbf{14.26} & 37.91 & 28.39 & \textbf{32.12} & 5.73 & 4.82 & \textbf{5.18} & 13.35 & 10.73 & \textbf{11.80} \\
    CoreRank Submodular             & 38.95 & 31.49 & 34.38 & 7.85 & 6.81 & 7.20 & 14.08 & 13.55 & 13.61 & 37.31 & 29.51 & 32.45 & 5.59 & 5.05 & 5.24 & 13.19 & 11.08 & 11.87 \\
    PageRank Submodular             & 40.58 & 32.87 & 35.86 & 9.20 & 7.77 & \textbf{8.32} & 15.59 & 14.14 & 14.64 & 37.72 & 28.86 & 32.35 & 6.35 & 5.46 & 5.82 & 13.35 & 11.60 & 12.30 \\
    TextRank                        & 39.47 & 32.57 & 35.19 & 7.74 & 6.62 & 7.05 & 14.80 & 13.03 & 13.69 & 37.60 & 28.79 & 32.32 & 6.63 & 5.53 & \textbf{5.98} & 14.18 & 11.18 & 12.41 \\
    ClusterRank                     & 38.32 & 31.51 & 34.10 & 6.93 & 5.95 & 6.31 & 13.69 & 12.40 & 12.84 & 35.66 & 26.58 & 30.14 & 4.53 & 3.99 & 4.21 & 12.10 & 9.71 & 10.69 \\
    Longest Greedy                  & 36.73 & 30.39 & 32.78 & 5.52 & 4.58 & 4.93 & 13.52 & 10.91 & 11.93 & 37.15 & 28.21 & 31.76 & 5.50 & 4.60 & 4.98 & 13.59 & 10.03 & 11.46 \\
    Random                          & 39.29 & 32.38 & 35.01 & 7.14 & 6.16 & 6.52 & 14.16 & 12.95 & 13.35 & 37.48 & 28.10 & 31.80 & 5.41 & 4.65 & 4.95 & 12.97 & 10.67 & 11.61 \\
\hline
\end{tabular}
}
\caption{Macro-averaged results for 350 and 450 word summaries (manual transcriptions). }
\label{table:res_manual}
\end{table*}

\section{Example Summaries}\label{app:examples}
Examples were generated from the manual transcriptions of meeting AMI TS3003c. Note that our system can also be interactively tested at \small\url{http://datascience.open-paas.org/abs_summ_app}.\\

\mdfdefinestyle{theoremstyle}{%
linecolor=gray!20,linewidth=1pt,%
leftline=false,
rightline=false,
topline=false,
bottomline=false,
frametitlerule=false,%
frametitlebackgroundcolor=gray!20,
innertopmargin=\topskip
}
\fontsize{11pt}{11pt}\selectfont
\begin{mdframed}[style=theoremstyle, frametitle={Reference Summary (254 words)}]
The project manager opened the meeting and recapped the decisions made in the previous meeting.
\\ The marketing expert discussed his personal preferences for the design of the remote and presented the results of trend-watching reports, which indicated that there is a need for products which are fancy, innovative, easy to use, in dark colors, in recognizable shapes, and in a familiar material like wood.
\\ The user interface designer discussed the option to include speech recognition and which functions to include on the remote.
\\ The industrial designer discussed which options he preferred for the remote in terms of energy sources, casing, case supplements, buttons, and chips.
\\ The team then discussed and made decisions regarding energy sources, speech recognition, LCD screens, chips, case materials and colors, case shape and orientation, and button orientation.
\\ The team members will look at the corporate website.
\\ The user interface designer will continue with what he has been working on.
\\ The industrial designer and user interface designer will work together.
\\ The remote will have a docking station.
\\ The remote will use a conventional battery and a docking station which recharges the battery.
\\ The remote will use an advanced chip.
\\ The remote will have changeable case covers.
\\ The case covers will be available in wood or plastic.
\\ The case will be single curved.
\\ Whether to use kinetic energy or a conventional battery with a docking station which recharges the remote.
\\ Whether to implement an LCD screen on the remote.
\\ Choosing between an LCD screen or speech recognition.
\\ Using wood for the case.
\end{mdframed}

\begin{mdframed}[style=theoremstyle, frametitle={Our System (250 words)}]
attract elderly people can use the remote control
\\ changing channels button on the right side that would certainly yield great options for the design of the remote
\\ personally i dont think that older people like to shake your remote control
\\ imagine that the remote control and the docking station
\\ remote control have to lay in your hand and right hand users
\\ finding an attractive way to control the remote control
\\ casing the manufacturing department can deliver a flat casing single or double curved casing
\\ top of that the lcd screen would help in making the remote control easier
\\ increase the price for which were selling our remote control
\\ remote controls are using a onoff button still on the top
\\ apply remote control on which you can apply different case covers
\\ button on your docking station which you can push and then it starts beeping
\\ surveys have indicated that especially wood is the material for older people
\\ mobile phones so like the nokia mobile phones when you can change the case
\\ greyblack colour for people prefer dark colours
\\ brings us to the discussion about our concepts
\\ docking station and small screen would be our main points of interest
\\ industrial designer and user interface designer are going to work
\\ innovativeness was about half of half as important as the fancy design
\\ efficient and cheaper to put it in the docking station
\\ case supplement and the buttons it really depends on the designer
\\ start by choosing a case
\\ deployed some trendwatchers to milan
\end{mdframed}

\begin{mdframed}[style=theoremstyle, frametitle={Our System (Baseline) (250 words)}]
apply remote controls on which you can apply different case for his remote control
\\ changing channels and changing volume button on both sides that would certainly yield great options for the design of the remote
\\ personally i dont think that older people like to shake their remote control
\\ finding an attractive way to control the remote control the i found some something about speech recognition
\\ imagine that the remote control and the docking station should be telephoneshaped
\\ casing the manufacturing department can deliver a flat casing single or double curved casing
\\ remote control have to lay in your hand and right hand users
\\ remote controls are using a onoff button over in this corner
\\ woodlike for the more exclusive people can use the remote control
\\ heard our industrial designer talk about flat single curved and double curved
\\ innovativeness this means functions which are not featured in other remote control
\\ button on your docking station which you can push and then it starts beeping
\\ greyblack colour for people prefer dark colours
\\ docking station and small screen would be our main points of interest
\\ special button for subtitles for people which c f who cant read small subtitles
\\ pretty big influence on production price and image unless we would start two product lines
\\ surveys have indicated that especially wood is the material for older people
\\ mobile phones so like the nokia mobile phones when you can change the case
\\ case the supplement and the buttons it really depends on the designer
\\ buttons
\end{mdframed}

\begin{mdframed}[style=theoremstyle, frametitle={Our System (KeyRank) (250 words)}]
changing case covers
\\ prefer a design where the remote control and the docking station
\\ greyblack colour for people prefer dark colours
\\ remote controls are using a onoff button over in this corner
\\ requirements are teletext docking station and small screen with some extras that button information
\\ apply remote controls on which you can apply different case covers
\\ woodlike for the more exclusive people can use the remote control
\\ casing the manufacturing department can deliver a flat casing single or double curved casing
\\ remote control have to lay in your hand and right hand users
\\ asked if w they would if people would pay more for speech recognition function would not make the remote control
\\ start by choosing a case
\\ innovativeness this means functions which are not featured in other remote controls
\\ top of that the lcd screen would help in making the remote control easier
\\ changing channels and changing volume button on both sides that would certainly yield great options for the design of the remote
\\ personally i dont think that older remotes are flat board smartboard
\\ button on your docking station which you can push and then it starts beeping
\\ case supplement and the buttons it really depends on the designer
\\ surveys have indicated that especially wood is the material for older people will recognise the button
\\ speak speech recognition and a special button for subtitles for people which c f who cant read small subtitles
\\ innovativeness was about half as important as the fancy design
\\ pretty big influence
\end{mdframed}

\begin{mdframed}[style=theoremstyle, frametitle={Our System (FluCovRank) (250 words)}]
elderly people can use the remote control
\\ remote controls are using a onoff button still on the top
\\ general idea of the concepts and the material for older people like to shake your remote control
\\ docking station and small screen would be our main points of interest
\\ industrial designer and user interface designer are going to work
\\ casing the manufacturing department can deliver single curved
\\ changing channels and changing volume button on both side that would certainly yield great options for the design of the remote
\\ button on your docking station which you can push and then it starts beeping
\\ imagine that the remote control will be standing up straight in the docking station will help them give the remote
\\ asked if w they would if people would pay more for speech recognition in a remote control you can call it and it gives an sig signal
\\ research about bi large lcd sh display for for displaying the the functions of the buttons
\\ case the supplement and the buttons it really depends on the designer
\\ pointed out earlier that a lot of remotes rsi
\\ innovativeness was about half of half as important as the fancy design
\\ push on the button for subtitles for people which c f who cant read small subtitles
\\ efficient and cheaper to put it in the docking station could be one of the marketing issues
\\ difficult to handle and to get in the right shape to older people
\\ talk about the energy source is rather fancy
\end{mdframed}

\end{document}